\documentclass[final]{cvpr}

\usepackage{nopageno}
\usepackage{times}
\usepackage{epsfig}
\usepackage{graphicx}
\usepackage{amsmath}
\usepackage{amssymb}
\usepackage{multirow}
\usepackage{booktabs}
\usepackage{soul}

\usepackage{algorithm}
\usepackage{algorithmic}
\usepackage{xcolor}
\usepackage{threeparttable}
\usepackage{textcomp}
\usepackage{subcaption}
\usepackage{mdwlist}
\usepackage{makecell}
\usepackage{stmaryrd}

\usepackage[pagebackref=true,breaklinks=true,colorlinks,bookmarks=false]{hyperref}

\def\cvprPaperID{531} % *** Enter the CVPR Paper ID here
\def\confYear{CVPR 2021}
\newcommand{\fu}{B}

\newcommand{\textapprox}{\raisebox{0.5ex}{\texttildelow}}

\newcommand*{\affmark}[1][*]{\textsuperscript{#1}}

\begin{document}

%%%%%%%%% TITLE
\title{Teachers Do More Than Teach: Compressing Image-to-Image Models}

\author{Qing Jin\affmark[1]\thanks{Work done while at Snap Inc.}\quad\quad
\quad Jian Ren\affmark[2]\quad\quad Oliver J.~Woodford\footnotemark[1] \quad\quad
Jiazhuo Wang\affmark[2]\quad\quad \\ Geng Yuan\affmark[1]\quad\quad Yanzhi Wang\affmark[1] \quad\quad Sergey Tulyakov\affmark[2]
\\
{\affmark[1]Northeastern University, USA\quad\quad\affmark[2]Snap Inc. }
}

\maketitle

\begin{abstract}
% ======== Final Version =========
Generative Adversarial Networks (GANs) have achieved huge success in generating high-fidelity images, however, they suffer from low efficiency due to tremendous computational cost and bulky memory usage. Recent efforts on compression GANs show noticeable progress in obtaining smaller generators by sacrificing image quality or involving a time-consuming searching process. In this work, we aim to address these issues by introducing a teacher network that provides a search space in which efficient network architectures can be found, in addition to performing knowledge distillation. First, we revisit the search space of generative models, introducing an inception-based residual block into generators. Second, to achieve target computation cost, we propose a one-step pruning algorithm that searches a student architecture from the teacher model and substantially reduces searching cost. It requires no $\ell^1$ sparsity regularization and its associated hyper-parameters, simplifying the training procedure. Finally, we propose to distill knowledge through maximizing feature similarity between teacher and student via Kernel Alignment (KA). Our compressed networks achieve similar or even better image fidelity (FID, mIoU) than the original models with much-reduced computational cost, e.g., MACs. Code will be released at~\href{https://github.com/snap-research/CAT}{https://github.com/snap-research/CAT}.
% ======== Final Version =========
\end{abstract}
\vspace{-20pt}
\section{Introduction}
% ======= Final Version =======%
\begin{figure}[t]
\centering
\includegraphics[width=\linewidth]{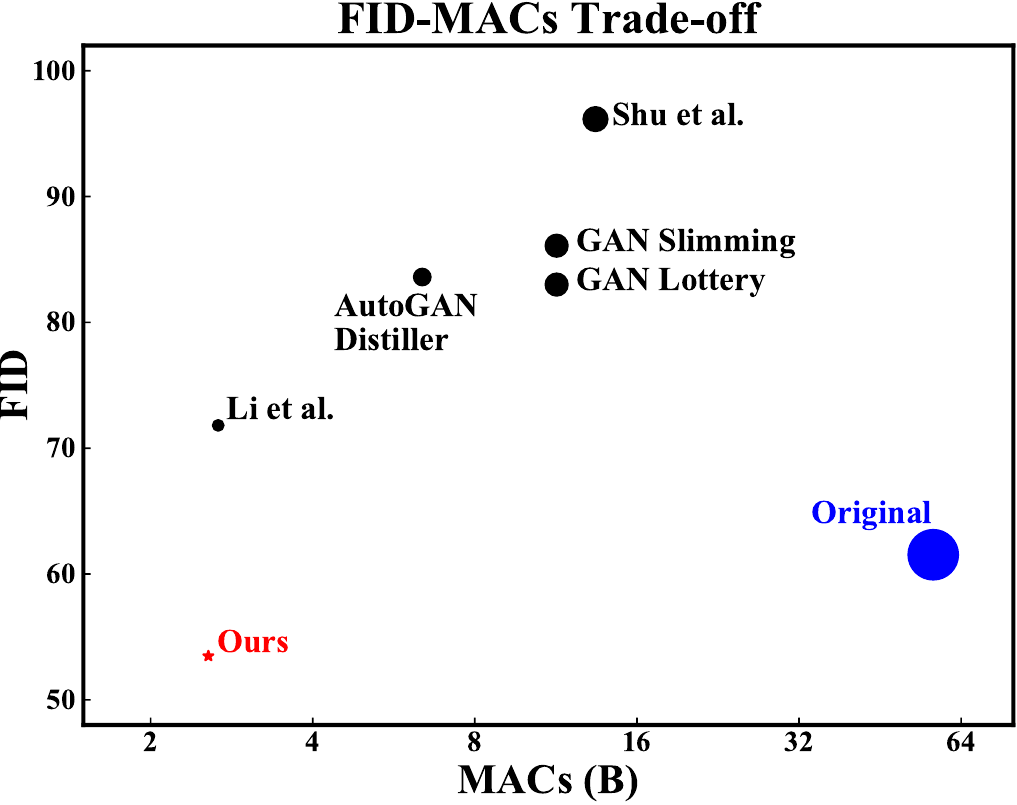}
\caption{Performance comparison between our and existing GAN compression techniques~\cite{chen2021gans,fu2020autogan,li2020gan,shu2019co,wang2020ganslimming} on CycleGAN~\cite{CycleGAN2017} for  Horse$\shortrightarrow$Zebra dataset. \emph{Smaller} MACs indicates more efficient models. \emph{Lower} FID indicates models can generate more realistic images.
Our method (red star) achieves the state-of-the-art performance-efficiency trade-off as it has the lowest FID with the smallest MACs.
}\label{fig:comp}
\vspace{-10pt}
\end{figure}
Generative adversarial networks (GANs), which synthesize images by adversarial training~\cite{goodfellow2014generative}, have witnessed tremendous progress in generating high-quality, high-resolution, and photo-realistic images and videos~\cite{brock2018large,karras2020analyzing,tian2021a}. In  conditional setting~\cite{mirza2014conditional}, the generation process is controlled via additional input signals, such as segmentation information~\cite{chai2020neural,park2019semantic,ren2020human,wang2018video,wang2018pix2pixHD}, class labels~\cite{zhang2019self}, and sketches~\cite{pix2pix2017,CycleGAN2017}. 
These techniques have seen applications in commercial image editing tools.
However, due to their massive computation complexity and bulky size, applying generative models at scale is less practical, especially on resource-constrained platforms, where low memory footprint, power consumption, and real-time execution are as, and often more, important than performance~\cite{li2020gan}.

To accelerate inference and save storage space for huge models without sacrificing performance, previous works propose to compress models with techniques including weight pruning~\cite{han2015deep}, channel slimming~\cite{liu2017learning,liu2018rethinking}, layer skipping~\cite{bolukbasi2017adaptive,wang2018skipnet}, patterned or block pruning~\cite{ding2018structured,li2019admm,liu2019autoslim,liu2020cocopie,ma2020pconv,ma2019non,ma2020image,ma2019resnet,niu2020patdnn,niu201926ms,yuan2019ultra,zhang2021structadmm}, and network quantization~\cite{choi2018pact,ding2019req,jin2019towards,jin2020adabits,jinneural,lin2019toward,yang2020fracbits}.
Specifically, these studies elaborate on compressing discriminative models for image classification, detection, or segmentation tasks.
The problem of compressing generative models, on the other hand, is less investigated, despite that typical generators are bulky in memory usage and inefficient during inference.
Up till now, only a handful of attempts exist~\cite{fu2020autogan,li2020gan,shu2019co,wang2020ganslimming}, all of which degenerate the quality of synthetic images compared to the original model (Fig.~\ref{fig:comp}). 

In this work, we focus on compressing image-to-image translation networks, such as CycleGAN~\cite{CycleGAN2017} and GauGAN~\cite{park2019semantic}. Existing compression method~\cite{li2020gan} obtains an efficient student model and employs two additional networks: teacher and supernet, where the former is for knowledge distillation and the latter for architecture search.
However, we argue that the supernet is not necessary, as the teacher can play its role.
Specifically, in our proposed framework, the teacher does more than teaching the student (\ie knowledge distillation)---it plays a central role in all aspects of the framework through three key contributions:
\begin{enumerate*}
\item We introduce a new network design that can be applied to both encoder-decoder architectures such as Pix2pix~\cite{pix2pix2017}, and decoder-style networks such as GauGAN~\cite{park2019semantic}. It serves as both the teacher network design, \emph{and} the architecture search space of the student.
\item
We directly prune the trained \emph{teacher} network using an efficient, one-step technique that removes certain channels in its generators to achieve a target computation budget,~\eg, the number of Multiply-Accumulate Operations (MACs).
This reduces architecture search costs by at least $10,000\times$ than the state-of-the-art compression method for generative models. Furthermore,  our pruning method only involves one hyper-parameter, making its application straightforward.
\item
We introduce a knowledge distillation technique based on the similarity between teacher and student models' feature spaces, which we call knowledge distillation with kernel alignment (\emph{KDKA}). 
KDKA directly forces feature representations from the two models to be similar, and avoids extra learnable layers~\cite{li2020gan} to match the different dimensions of teacher and student feature spaces, which could otherwise lead to information leakage.
\end{enumerate*}
We name our method as \textbf{CAT} as we show teacher model can and should do \textbf{C}ompression \textbf{A}nd \textbf{T}eaching (distillation) jointly, which we find is beneficial for finding generative networks with smaller MACs, using much lower computational resource than prior work. More importantly, our compressed networks can achieve similar or even better performance than their original counterparts (Tab.~\ref{tab:comparison}).
\section{Related Work}
%% ============= final version ================ %%
Due to their high computation cost, running GANs on resource-constrained devices in real-time remains a challenging problem. As a result, GAN compression has garnered attention recently. Existing methods~\cite{aguinaldo2019compressing,chen2020distilling,fu2020autogan,li2020gan,shu2019co,wang2020ganslimming} exploit network architecture search/pruning and knowledge distillation (discussed below).
Although they can compress the original models (\eg, CycleGAN~\cite{CycleGAN2017}) to a relatively small MACs, all these methods suffers from sacrifice on performance. In contrast, our method finds smaller networks than existing compressed GAN models, whilst \emph{improves} performance over the original models, such as Pix2pix~\cite{pix2pix2017}, CycleGAN~\cite{CycleGAN2017}, and GauGAN~\cite{park2019semantic}.

\noindent\textbf{Network architecture search \& pruning.} 
To determine the structure of a pruned model, previous work employs neural architecture search (NAS)~\cite{cai2018proxylessnas,chen2019progressive,li2020neural,liu2019autoslim,liu2018darts,liu2020autocompress,liu2020cocopie,lu2020nsganetv2,lu2020neural,mei2019atomnas,ren2019eigen,tan2019mnasnet,wu2019fbnet,yu2020bignas,zoph2016neural} and pruning techniques~\cite{bolukbasi2017adaptive,ding2017circnn,ding2018structured,li2019admm,liu2020cocopie,liu2017learning,liu2018rethinking,ma2020pconv,ma2019non,ma2020image,ma2019resnet,niu2020patdnn,niu201926ms,shi2020csb,wang2018skipnet,yu2019autoslim,yu2019universally,yu2020cakes,yuan2019ultra,zhang2021structadmm}, where the number of channels and/or operations can be optimized automatically. 
Applying these methods directly on generative models can lead to inferior performance of compressed models than their original counterparts.
For example, Shu~\etal~\cite{shu2019co} employ an evolutionary algorithm~\cite{real2017large} and 
Fu~\etal~\cite{fu2020autogan} engage differentiable network design~\cite{liu2018darts}, while Li~\etal~\cite{li2020gan} train a supernet with random sampling technique~\cite{cai2019once,guo2020single,yu2019universally,yu2020bignas} to select the optimal architecture. The common key drawback of these methods is the slow searching process. In contrast, directly pruning on a pre-trained model is much faster.
Following previous methods of network slimming~\cite{liu2017learning,liu2018rethinking}, Wang~\etal~\cite{wang2020ganslimming} apply $\ell^1$ regularization to generative models for channel pruning. However, they report performance degradation compared to the original network. 
Besides, these pruning methods require tuning additional hyper-parameters for $\ell^1$ regularization to encourage channel-wise sparsity~\cite{liu2017learning,liu2018rethinking} and even more hyper-parameters to decide the number of channels to be pruned~\cite{mei2019atomnas}, making the process tedious.
Additionally, GAN training involves optimizing multiple objective functions, and the associated hyper-parameters make the training process even harder.
Recently, lottery ticket hypothesis~\cite{frankle2018lottery} is also investigated on GAN problem~\cite{chen2021gans}, while the performance is not satisfactory.

\noindent\textbf{Knowledge distillation}~\cite{hinton2015distilling} is a technique to transfer knowledge from a larger, teacher network to a smaller, student network, and has been used for model compression in various computer vision tasks~\cite{chen2017learning,chen2020distilling,luo2016face,yim2017gift,lopez2015unifying}. A recent survey~\cite{gou2020knowledge} categorizes knowledge distillation as response-based, feature-based, or relation-based. Most GAN compression methods~\cite{aguinaldo2019compressing,chen2020distilling,fu2020autogan} use response-based distillation, enforcing the synthesized images from the teacher and student networks to be the same. Li \etal~\cite{li2020gan} apply feature-based distillation by introducing extra layers to match feature sizes between the teacher and student, and minimizing the differences of these embeddings using mean squared error (MSE) loss. 
However, this has the potential problem that some information can be stored in those extra layers, without being passed on to the student. Here, we propose to distill knowledge by directly maximizing the similarity between features from teacher and student models.
\section{Methods}

% ======= final version ======= %
In this section, we show our method for searching a compressed student generator from a teacher generator. 
We revisit the network design of conditional image generation models and introduce inception-based residual blocks (Sec.~\ref{sec:search_space}). 
The teacher model is built upon the proposed block design and can serve two purposes. 
First, we show that the teacher model can be viewed as a large search space that enables \emph{one-shot} neural architecture search without training an extra supernet.
With the proposed one-step pruning method, a computationally efficient network that satisfies a given computational budget can be found instantly (Sec.~\ref{sec:searching}). 
Second, we show the teacher model itself is sufficient for knowledge distillation, without necessity of introducing extra layers.
By maximizing the similarity between intermediate features of teacher and student network directly, where features of the two networks contain different numbers of channels, we can effectively transfer knowledge from teacher to student (Sec.~\ref{sec:distillation}).

\begin{figure}[t]
\centering
\includegraphics[width=1\linewidth]{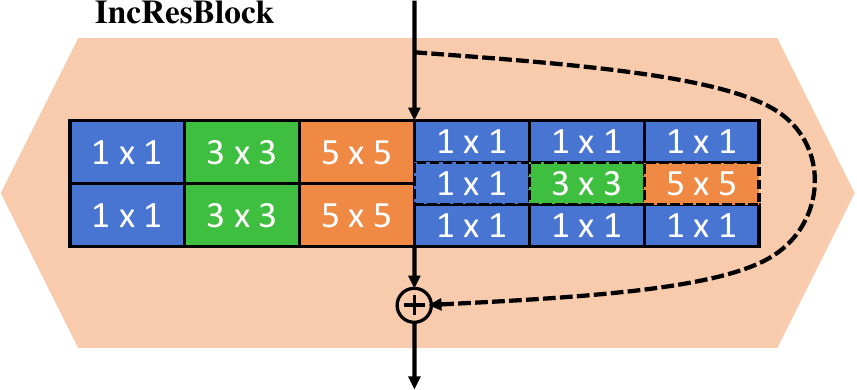}
\caption{
IncResBlock includes three conventional convolution blocks and three depth-wise convolution blocks (dashed border), both with kernels sizes of 1, 3, 5. Normalization layers (\eg, BN), and ReLU, are applied between each two consecutive convolution layers.
A normalization layer that can be inserted after summing features from the six blocks and the residual connection are optional. Unless otherwise stated, both are applied by default.}\label{fig:block}
\end{figure}
\subsection{Design of Teacher Generator}\label{sec:search_space}
Existing efforts leverage supernet to introduce search space that contains more efficient networks~\cite{cai2019once,guo2020single,li2020gan}. 
The optimization of supernet can lead to extra training costs.
However, as we already have a teacher network in hand, searching efficient student from the teacher model should be more straightforward, as long as the teacher network contains a large searching space. 
In this way, the teacher network can perform both knowledge distillation and provide search space.
Therefore, the goal of obtaining a good supernet can be changed to design a teacher generator that can synthesize high fidelity images; and itself contains a reasonable search space.

\noindent\textbf{Inception-based residual block.}
With the above goal bearing in mind, we design a new architecture for the image generation tasks so that a pre-trained teacher generator with such architecture can serve as a large search space. 
We aim to search for a smaller student network that can have different operations (\eg, convolution layers with various kernel size) and different numbers of channels than the teacher network through pruning. 
Towards this end, we adopt the widely used inception module on discriminative models~\cite{mei2019atomnas,szegedy2016inception,zoph2018learning} to the image generators and propose the inception-based residual block (IncResBlock). 
A conventional residual block in generators only contains convolution layers with one kernel size (\eg, $3\times3$), while in IncResBlock, as shown in Fig.~\ref{fig:block}, we introduce convolution layers with different kernel sizes, including $1\times1$, $3\times3$, and $5\times5$. Additionally, we incorporate depth-wise blocks~\cite{howard2017mobilenets} into IncResBlock as depth-wise convolution layers typically require less computation cost without sacrificing the performance, and are particularly suitable for models deployed on mobile devices~\cite{sandler2018mobilenetv2}.
Specifically, the IncResBlock includes six types of operations, with two types of convolution layers and three different kernel-sizes.
To achieve similar total computation cost, we set the number of output channels for the first convolution layers of each operations to that of the original residual blocks divided by 
six, which is the number of different operations in the IncResBlock. We find the performance is maintained thanks to the architecture design. 

To get our teacher networks, for Pix2pix and CycleGAN, we replace all residual blocks in original models with the IncResBlock. 
For GauGAN, we apply IncResBlock in both the SPADE modules and the residual blocks.
More details are illustrated in the supplementary materials.

\subsection{Search from Teacher Generator via Pruning}\label{sec:searching}

With the teacher network introduced, we search a compressed student network from it. 
Our searching algorithm includes two parts. The first one is deciding a threshold based on the given computational budget, and the second one is pruning channels with a scale less than a threshold. Compared with existing iterative pruning methods~\cite{liu2017learning,mei2019atomnas}, we only perform pruning once, and we name our searching algorithm as \emph{one-step pruning}.

\noindent\textbf{Automatically threshold searching.} 
Following existing efforts~\cite{liu2017learning,liu2018rethinking},
we prune the channels through the magnitudes of scaling factors in
normalization layers,  such as Batch Normalization (BN)~\cite{ioffe2015batch} and Instance Normalization (IN)~\cite{ulyanov2016instance}. To this end, a threshold is required to choose channels to prune.
As we train the teacher model without regularization, there is no constraint to force the teacher model to be sparse. The magnitude of scaling factors from the normalization layers is not guaranteed to be small. 
Thus, the previous iterative pruning methods, which remove channels using a manually designed threshold, are not suitable for our network.

To solve this, we determine the threshold by a given computation budget, which can be MACs or latency. All channels with scale smaller than the threshold are pruned until the final model achieves the target computation budget. We find the scale threshold by binary search on the scaling factors of normalization layers from the pre-trained teacher model. 
Specifically, we temporarily prune all channels with a scaling factor magnitude smaller than the threshold and measure the computational cost of the pruned model. If it is smaller than the budget, the model is pruned too much and we search in the lower interval to get a smaller threshold; otherwise, we search in the upper interval to get a larger value. During this process, we also keep the 
number of output channels for convolution layers outside the IncResBlock larger than a pre-defined value to avoid an invalid model. Details of the algorithm are illustrated in Algorithm~\ref{alg:prune}.

\noindent\textbf{Channel pruning.}
With the threshold decided, we perform network searching via pruning. 
Given an IncResBlock, it is possible to change both the number of channels in each layer and modify the operation, such that,~\eg, one IncResBlock may only include layers with kernel sizes $1\times1$ and $3\times3$.
Similar to Mei~\etal~\cite{mei2019atomnas}, we prune channels of the normalization layers together with the corresponding convolution layers.
Specifically, we prune the first normalization layers for each operation in IncResBlock, namely the ones after the first $k\times k$ convolution layers for conventional operations and the ones after the first $1\times 1$ convolution layers for depth-wise operations.

\begin{algorithm}[htb!]
\caption{{Searching via One-Step Pruning.} }
\label{alg:prune}
\begin{algorithmic}[1]
\REQUIRE Computational budget $T_{\mathrm{b}}$, teacher model $\mathrm{G}_\mathrm{T}$, scaling factors $\gamma_i^{(l)}$ (used for pruning) of the $i$-th channel in normalization layers $N^{(l)}$$\in$$\mathrm{G}_\mathrm{T}$, minimum \# output channels $c_{lb}$ for convolution layers (outside the IncResBlock).
\\
\ENSURE pruned student architecture 
$\mathrm{G}_\mathrm{S}$.\\
\STATE Initialize scale lower bound  $\gamma_{lo}$: 
$\gamma_{lo}\leftarrow\min\limits_{i,l}\lvert\gamma_i^{(l)}\rvert$.\\
\STATE Initialize scale upper bound  $\gamma_{hi}$: 
$\gamma_{hi}\leftarrow\max\limits_{i,l}\lvert\gamma_i^{(l)}\rvert$. \\
\WHILE{$\gamma_{lo}<\gamma_{hi}$}
\STATE$\gamma_{th}\leftarrow(\gamma_{lo}+\gamma_{hi})/2$
\STATE Prune channels satisfying $\lvert\gamma_i^{(l)}\rvert<\gamma_{th}$ on $\mathrm{G}_\mathrm{T}$ while keep $c_{lb}$
to get $\mathrm{G}_\mathrm{S}$
\STATE$T\leftarrow$ computational cost of $\mathrm{G}_\mathrm{S}$
\IF{$T>T_{\mathrm{b}}$}
\STATE $\gamma_{lo}\leftarrow\gamma_{th}$
\ELSE
\STATE $\gamma_{hi}\leftarrow\gamma_{th}$
\ENDIF
\ENDWHILE
\end{algorithmic}
\end{algorithm}

\noindent\textbf{Discussion.}
Our searching algorithm is different from previous works that focus on compressing generative models in the following three perspectives. 
First, we search an efficient network from a \emph{pre-trained} teacher model without utilizing an extra supernet~\cite{li2020gan}. 
Second, we show the scales of the normalization layers in the \emph{pre-trained} teacher network are sufficient for pruning, therefore, weight regularization for iterative pruning~\cite{mei2019atomnas,wang2020ganslimming} might not be necessary for the generation tasks. 
Third, the teacher network can be compressed to several different architectures, and we can find the student network that satisfies an arbitrary type of computational cost, \eg, MACs, under any value of predefined budget during the searching directly. 
Such differences bring us three advantages. 
First, searching cost is significantly reduced without introducing extra network. 
Second, removing the weight regularization, \eg, $\ell^1$-norm, eases the searching process as a bunch of hyper-parameters are reduced, which we find are hard to tune in practice. 
Third, we have more flexibility to choose a student network with required computational cost.

\subsection{Distillation from Teacher Generator}\label{sec:distillation}
\begin{figure}[]
\centering
\begin{subfigure}[t]{.28\textwidth}
\centering
\includegraphics[width=.8\linewidth]{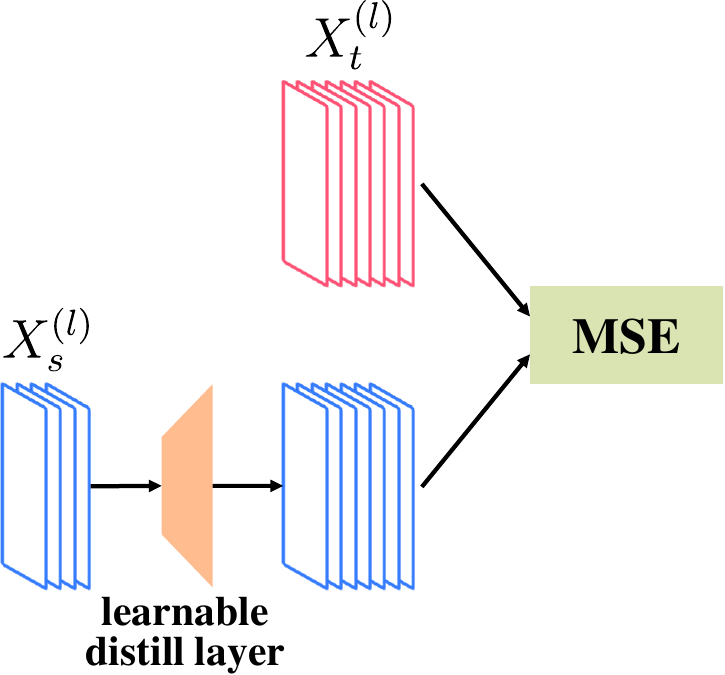}\label{fig:mse}
\end{subfigure}
\hspace{-2ex}
\begin{subfigure}[t]{.2\textwidth}
\centering
\includegraphics[width=.8\linewidth]{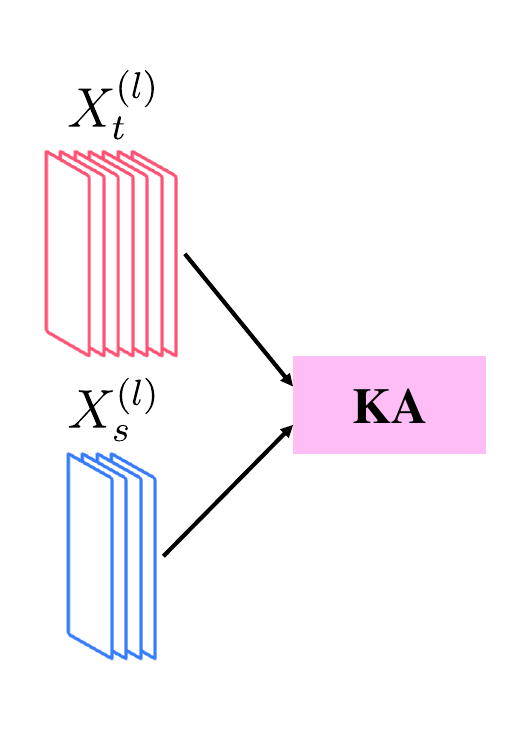}\label{fig:gcka}
\end{subfigure}
\caption{
\textbf{Left}: Knowledge distillation with MSE loss requires extra learnable layers to map features into the same number of channels. \textbf{Right}: Our proposed KDKA maximizes similarity of KA between features directly.}\label{fig:distill}
\end{figure}

After obtaining a student network architecture, we train it from scratch, leveraging the teacher model for knowledge distillation. 
In particular, we transfer knowledge between the two networks' \emph{feature} spaces, since this has been shown~\cite{li2020gan} to achieve better performance than reconstructing images synthesized by the teacher~\cite{fu2020autogan}.
With different numbers of channels between teacher and student layers, Li~\etal~\cite{li2020gan} introduce auxiliary, learnable layers that project the student features into the same dimensional space as the teacher, as shown in Fig.~\ref{fig:distill}. Whilst equalizing the number of channels between the two networks, these layers can also impact the efficacy of distillation, since some information can be stored in these extra layers.
To avoid information loss, we propose to encourage similarity between the two feature spaces directly.

\subsubsection{Similarity-based Knowledge Distillation}
We develop our distillation method based on centered kernel alignment (CKA)~\cite{cortes2012algorithms,cristianini2006kernel}, a similarity index between two matrices, $X\in\mathbb{R}^{n\times p_1}$ and $Y\in\mathbb{R}^{n\times p_2}$, where after centering the kernel alignment (KA) is calculated, which is defined as\footnote{The identity $\Vert Y^\mathrm{T}X\Vert_\mathrm{F}^2=\langle\mathrm{vec}(XX^\mathrm{T}),\mathrm{vec}(YY^\mathrm{T})\rangle$ is used to achieve computational complexity of $\mathcal{O}(n^2hw\max(c_1,c_2))$~\cite{kornblith2019similarity}.}
\begin{equation}\label{eqn:cka}
    \mathrm{KA}(X,Y)=\frac{\Vert Y^\mathrm{T}X\Vert_\mathrm{F}^2}{\Vert X^\mathrm{T}X\Vert_\mathrm{F}\Vert Y^\mathrm{T}Y\Vert_\mathrm{F}}.
\end{equation}
It is invariant to an orthogonal transform and isotropic scaling of the rows, but is sensitive to an invertible linear transform. Importantly, $p_1$ and $p_2$ can differ.
Kornblith \etal~\cite{kornblith2019similarity} use this index to compute the similarity between different learned feature representations of varying lengths ($p_1=hwc_1$ \& $p_2=hwc_2$, where $h$, $w$ and $c_\cdot$ are the height, width and number of channels of the respective layer tensors; $n$ is the batch size).

\noindent\textbf{KDKA.}
To compare similarity between teacher and student features, we adopt KA of the two tensors $X$ and $Y$ without centering. We find that centering is not necessary for our purpose of similarity-based knowledge distillation. To perform distillation, we maximize the similarity between features of teacher and student networks by maximizing KA.

\noindent\subsubsection{Distillation Loss}
We conduct distillation on the feature space. Let  $\mathcal{S}_\mathrm{KD}$ denote the set of layers for performing knowledge distillation, whereas $X_t^{(l)}$ and $X_s^{(l)}$ denote feature tensors of layer $l$ from the teacher and student networks, respectively. We minimize the distillation loss $\mathcal{L}_\mathrm{dist}$ as follows:
\begin{equation}
    \mathcal{L}_\mathrm{dist}=-\sum_{l\in\mathcal{S}_\mathrm{KD}}\mathrm{KA}(X_t^{(l)}, X_s^{(l)}),
\end{equation}
where the minus sign is introduced as we intend to maximize feature similarity between student and teacher models.

\subsection{Learning}
We train teacher networks using the original loss functions, which includes an adversarial loss $\mathcal{L}_\mathrm{adv}$ as follows:
\begin{equation}\label{eqn:adv}
    \mathcal{L}_{\mathrm{adv}}= {\mathbb{E}_{\mathbf{x},\mathbf{y}}}\left [\log D(\mathbf{x}, \mathbf{y})\right ] + {\mathbb{E}_{\mathbf{x}}}\left[ \log (1 - D(\mathbf{x}, G(\mathbf{x}))) \right ],
\end{equation}
where $\mathbf{x}$ and $\mathbf{y}$ denote the input and real images, and $D$ and $G$ denote the discriminator and generator, respectively.

\noindent\textbf{Full objective for student.}
For the training of student generator for CycleGAN, we adopt the setting from~\cite{li2020gan} where we use the data generated from teacher network to form paired data and train the student the same way as Pix2pix with a reconstruction loss $\mathcal{L}_\mathrm{recon}$. Therefore, for CycleGAN and Pix2pix, the overall loss function for student training is:
\begin{equation}\label{eqn:pix2pix}
    \mathcal{L}_\mathrm{T} = \lambda_\mathrm{adv}\mathcal{L}_\mathrm{adv} + \lambda_\mathrm{recon} \mathcal{L}_\mathrm{recon} +  \lambda_\mathrm{dist}\mathcal{L}_\mathrm{dist}.
\end{equation}
For the training of GauGAN, there is an additional feature matching loss $\mathcal{L}_\mathrm{fm}$~\cite{wang2018pix2pixHD}, and the overall loss function is as follows:
\begin{equation}\label{eqn:spade}
    \mathcal{L}_\mathrm{T} = \lambda_\mathrm{adv}\mathcal{L}_\mathrm{adv} + \lambda_\mathrm{recon}\mathcal{L}_\mathrm{recon} +  \lambda_\mathrm{fm}\mathcal{L}_\mathrm{fm} + \lambda_\mathrm{dist}\mathcal{L}_\mathrm{dist}.
\end{equation}
$\lambda_\mathrm{adv}$, $\lambda_\mathrm{recon}$, $\lambda_\mathrm{dist}$ and  $\lambda_\mathrm{fm}$ in Eqn.~\ref{eqn:pix2pix} and Eqn.~\ref{eqn:spade} indicate the hyper-parameters that balance the losses.

% ======= final version ======= %
\begin{table*}[h]
    \centering
    \begin{threeparttable}
    \caption{Quantitative comparison between different compression techniques for Image-to-Image models.
    We use mIoU to evaluate the generation quality of Cityspaces and FID for other datasets.
    Higher mIoU or lower FID indicates better performance.
    % For Cityscapes, we use mIoU as the metric, where larger values indicate better image quality. For other datasets, we use FID and lower values are better.
    }
    \label{tab:comparison}
    \begin{tabular}{lclcccc}
        \toprule
        % \multirow{2}{*}{Model} & \multirow{2}{*}{Dataset} &\multirow{2}{*}{Method} & \multirow{2}{*}{MACs} & \multicolumn{2}{c}{Metric} & \multirow{2}{*}{Searching Cost}  \\ 
    % \cmidrule{5-6}
    % & & &  & FID$\downarrow$ & mIoU$\uparrow$ & \\\midrule
    Model & Dataset & Method & MACs & FID$\downarrow$ & mIoU$\uparrow$ \\ \midrule
        % numbers for cyclegan
        \multirow{7.8}{*}{CycleGAN} & \multirow{7.8}{*}{Horse$\shortrightarrow$Zebra} & Original~\cite{CycleGAN2017,li2020gan} & 56.8\fu &  61.53 & - \\ \cmidrule{3-6}
        & & Shu \textit{et al.}~\cite{shu2019co} & 13.4\fu  & 96.15 & - \\
        & & AutoGAN Distiller~\cite{fu2020autogan} & 6.39\fu & 83.60 & - \\
        & & GAN Slimming~\cite{wang2020ganslimming} & 11.25\fu & 86.09 & - \\
        & & GAN Lottery~\cite{chen2021gans} & \textapprox11.35\fu\tnote{$\dagger$} & \textapprox83.00\tnote{$\dagger$} & - \\
        & & Li \textit{et al.}~\cite{li2020gan} & 2.67\fu  &  71.81 & - \\
        % & & Ours w/ Atomnas &\textcolor{red}{xx}  & \textcolor{red}{xx} & - & \textcolor{red}{xx} \\
        & & \textbf{CAT (Ours)} & \textbf{2.56\fu}  &  \textbf{53.48} & - \\
        % \cmidrule{2-6}
        % & \multirow{3.5}{*}{Zebra$\shortrightarrow$Horse} & Original~\cite{CycleGAN2017,wang2020ganslimming} & 56.8\fu &  148.81 & - \\
        % \cmidrule{3-6}
        % & & GAN Slimming~\cite{wang2020ganslimming} & 11.81\fu & 120.01 & - \\
        % & & CAT (Ours) & 2.59\fu &  142.68 & - \\
        \midrule
        
        % ======= numbers for pix2pix edge to shoes
        % \multirow{12}{*}{Pix2pix} & \multirow{4}{*}{edge$\rightarrow$shoes}  & Original~\cite{pix2pix2017} &  56.8\fu  &   24.18 & - & -\\\cmidrule{3-7}
        % & & Li \textit{et al.}~\cite{li2020gan} & 4.81\fu &  26.60 & - & \textcolor{red}{xx\scu}\tnote{$\dagger$}\\
        % & & \textcolor{red}{XXX-A} (\textbf{Ours}) & \textcolor{red}{xx}  & \textcolor{red}{xx} & - & \textcolor{red}{xx\scu}\\\cmidrule{2-7}
        % =======  numbers for pix2pix Cityscapes
        \multirow{8}{*}{Pix2pix}  & \multirow{3}{*}{Cityscapes}  & Original~\cite{pix2pix2017,li2020gan} &  56.8\fu  & - & 42.06 \\\cmidrule{3-6}
        & & Li \textit{et al.}~\cite{li2020gan} &  5.66\fu & - & 40.77 \\
        % & & Ours w/ Atomnas & \textcolor{red}{xx}  &  - & \textcolor{red}{xx} & \textcolor{red}{xx} \\
        & & \textbf{CAT (Ours)} & \textbf{5.57\fu} & - & \textbf{42.65} \\\cmidrule{2-6}
        % ======= numbers for pix2pix map
        & \multirow{3}{*}{Map$\shortrightarrow$Aerial photo}  & Original~\cite{pix2pix2017,li2020gan} &  56.8\fu  &  47.76 & - \\\cmidrule{3-6}
        & & Li \textit{et al.}~\cite{li2020gan} &  4.68\fu  &  48.02 & - \\
        % & & Ours w/ Atomnas & \textcolor{red}{xx} & \textcolor{red}{xx} & - & \textcolor{red}{xx}\\
        & & \textbf{CAT (Ours)} & \textbf{4.59\fu}  &   \textbf{45.63} & - \\\midrule
        % ======= numbers for gaugan
        \multirow{4.5}{*}{GauGAN} & \multirow{4.5}{*}{Cityscapes}  & Original~\cite{park2019semantic,li2020gan} &  281\fu & - &62.18 \\\cmidrule{3-6}
        & & Li \textit{et al.}~\cite{li2020gan} & 31.7\fu  & - & 61.22 \\
        & &  \textbf{CAT-A (Ours)}  & \textbf{29.9\fu} & - & \textbf{62.35} \\
        % \cmidrule{3-6}
        %& & & 28.0\fu & - & 59.78 &\textcolor{red}{xx}\\
        %& & & 20.0\fu & - & 58.54 &\textcolor{red}{xx}\\
        %& & & 15.0\fu & - & 57.88 &\textcolor{red}{xx}\\
        %& & & 10.0\fu & - & 56.13 &\textcolor{red}{xx}\\
        & &  \textbf{CAT-B (Ours)} & 5.52\fu & - & 54.71
        % \\\midrule
        % % ======= numbers for clade
        % \multirow{4.5}{*}{Clade} & \multirow{4.5}{*}{Cityscapes}  & Original~\cite{tan2020rethinking} & 75.2\fu & - & 59.26\\
        % & & Clade-Avg (\textbf{Ours})  & 75.2\fu  & - & 58.79 \\\cmidrule{3-6}
        % %& & Clade-Avg (\textbf{Ours})  & 53.4\fu  & - & 60.97 \\
        % & & \textbf{CAT-A (Ours)}  & \textbf{29.4\fu}  & -& \textbf{61.93} \\
        % & & \textbf{CAT-B (Ours)}  & 5.53\fu  & - & 55.78
        \\\bottomrule
    \end{tabular}
    \begin{tablenotes}\footnotesize
      \item[$\dagger$]
      Estimated from Fig.~11 in~\cite{chen2021gans}.
    \end{tablenotes}
    \end{threeparttable}
\end{table*}

\begin{table}[h]
    \centering
    \caption{Further quantitative comparison on KID between different compression techniques for Image-to-Image models, where lower KID indicates better performance.
    }
    \label{tab:comparison_kid}
    \scalebox{0.8}{
    \begin{tabular}{ccccc}
        \toprule
        Model & Dataset & Method & MACs & KID$\downarrow$ \\ \midrule
        \multirow{2}{*}{CycleGAN} & \multirow{2}{*}{Horse$\shortrightarrow$Zebra} & Original~\cite{CycleGAN2017} & 56.8\fu & 0.020$\pm$0.002\\
        & & \textbf{CAT (Ours)} & \textbf{2.56\fu} & \textbf{0.015$\pm$0.001}\\
        % \cmidrule{2-5}
        % & \multirow{2}{*}{Zebra$\shortrightarrow$Horse} & Original~\cite{CycleGAN2017} & 56.8\fu & 0.030$\pm$0.002\\
        % & & CAT (Ours) & 2.59\fu & 0.036$\pm$0.002 \\
        \midrule
        \multirow{2}{*}{Pix2pix} & \multirow{2}{*}{Map$\shortrightarrow$Aerial} & Original~\cite{pix2pix2017} & 56.8\fu & 0.154$\pm$0.010\\
        & & \textbf{CAT (Ours)} & \textbf{4.59\fu} & \textbf{0.012$\pm$0.002} \\
        \midrule
        \multirow{3}{*}{GauGAN} & \multirow{3}{*}{Cityscapes} & Original~\cite{park2019semantic} & 281\fu & 0.026$\pm $0.003\\
        & & \textbf{CAT-A (Ours)} & \textbf{29.9\fu} & \textbf{0.014$\pm$0.002} \\
        & & \textbf{CAT-B (Ours)} & \textbf{5.5\fu} & \textbf{0.013$\pm$0.002} \\
        \bottomrule
    \end{tabular}
    }
    \vspace{-5mm}
\end{table}

% \begin{table}[h]
%     \centering
%     \caption{Further quantitative comparison on KID between different compression techniques for Image-to-Image models, where lower KID indicates better performance.
%     }
%     \label{tab:comparison_kid}
%     \scalebox{0.8}{
%     \begin{tabular}{ccccc}
%         \toprule
%         Model & Dataset & Method & MACs & KID$\downarrow$ \\ \midrule
%         \multirow{4.5}{*}{CycleGAN} & \multirow{2}{*}{Horse$\shortrightarrow$Zebra} & Original~\cite{CycleGAN2017} & 56.8\fu & 0.020$\pm$0.002\\
%         & & \textbf{CAT (Ours)} & \textbf{2.55\fu} & \textbf{0.017$\pm$0.002}\\
%         \cmidrule{2-5}
%         & \multirow{2}{*}{Zebra$\shortrightarrow$Horse} & Original~\cite{CycleGAN2017} & 56.8\fu & 0.030$\pm$0.002\\
%         & & CAT (Ours) & 2.59\fu & 0.036$\pm$0.002 \\
%         \midrule
%         \multirow{2}{*}{Pix2pix} & \multirow{2}{*}{Map$\shortrightarrow$Aerial} & Original~\cite{pix2pix2017} & 56.8\fu & 0.154$\pm$0.010\\
%         & & \textbf{CAT (Ours)} & \textbf{4.6\fu} & \textbf{0.009$\pm$0.002} \\
%         \midrule
%         \multirow{3}{*}{GauGAN} & \multirow{3}{*}{Cityscapes} & Original~\cite{park2019semantic} & 281\fu & 0.026$\pm $0.003\\
%         & & \textbf{CAT-A (Ours)} & \textbf{29.9\fu} & \textbf{0.014$\pm$0.002} \\
%         & & \textbf{CAT-B (Ours)} & \textbf{5.5\fu} & \textbf{0.013$\pm$0.002} \\
%         \bottomrule
%     \end{tabular}
%     }
%     \vspace{-5mm}
% \end{table}
\section{Experiments}

In this section, we show the results of compressing 
image-to-image models. We introduce more details about network training and  architectures, together with more qualitative results in the \emph{supplementary materials}.

\subsection{Basic Setting}
\noindent\textbf{Models.} 
We conduct experiments on generation models, including Pix2pix~\cite{pix2pix2017}, CycleGAN~\cite{CycleGAN2017}, and GauGAN~\cite{park2019semantic}. 
Following~\cite{li2020gan}, we inherit the teacher discriminator by using the same architecture and the pre-trained weights, and finetune it with the student generator for student training.

\begin{figure*}[]
\centering
\includegraphics[width=1\linewidth]{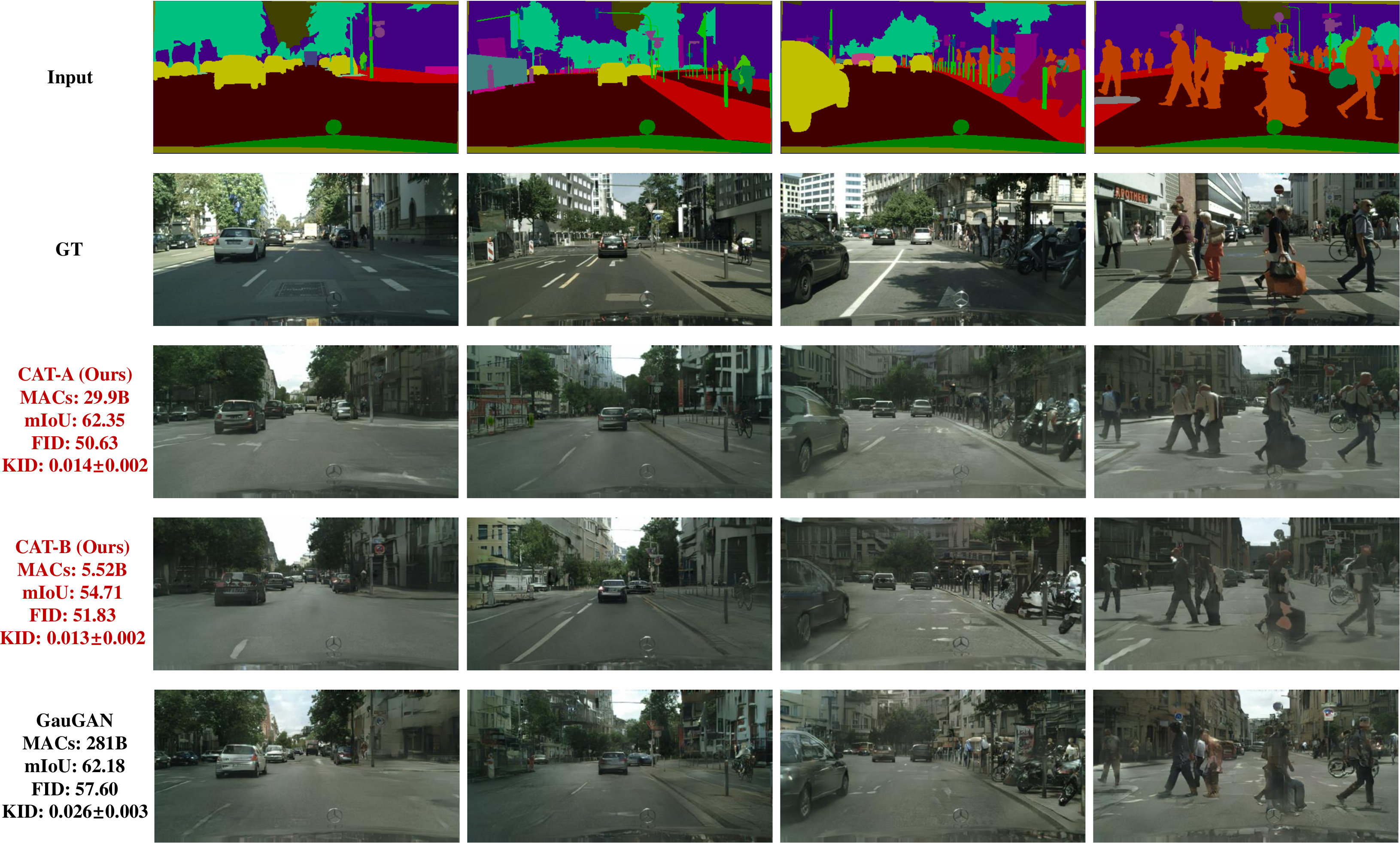}
\caption{
Qualitative results on Cityscapes dataset. Images generated by our compressed model (CAT-A, third row) have higher mIoU and lower FID than the original GauGAN model (fifth row), even with much reduced computational cost.
For our CAT-B model (fourth row, $50.9\times$ compressed than GauGAN), although it has lower mIoU, the CAT-B model can synthesize higher fidelity images (lower FID) than GauGAN.}\label{fig:city_gaugan}
\end{figure*}

\noindent\textbf{Datasets.} We examine our method on the following datasets. \emph{Horse$\shortrightarrow$Zebra} and \emph{Zebra$\shortrightarrow$Horse} are two datasets from CycleGAN~\cite{CycleGAN2017}, which converts horse images to zebra and vice versa. There are $1,187$ horse images and $1,474$ zebra images. \emph{Cityscapes}~\cite{cordts2016cityscapes} is a dataset for mapping semantic inputs to images of street scenes. There are $2,975$ training and $500$ validation data, and we apply Pix2pix and GauGAN models on it. \emph{Map$\shortrightarrow$Aerial photo} contains $2,194$ images~\cite{pix2pix2017}, and we apply Pix2pix model on it.

\noindent\textbf{Evaluation metrics.} 
We adopt two standard metrics for the evaluation of generative models.
For the Cityspaces dataset, we follow existing works~\cite{pix2pix2017,park2019semantic} to use a semantic segmentation metric to evaluate the quality of synthetic images. We run an image segmentation model, which is DRN-D-105~\cite{yu2017dilated}, on the generated images to calculate mean Intersection over Union (mIoU). A higher value of mIoU indicates better quality of generated images. For other datasets, we apply commonly used Fr$\mathrm{\acute{e}}$chet Inception Distance (FID)~\cite{heusel2017gans}, as it estimates the distribution between real and generated images. We also adopt a recent proposed metric named Kernel Inception Distance (KID)~\cite{binkowski2018demystifying} for more thorough comparison. A lower FID or KID value indicates better model performance. 

\subsection{Comparison Results}
\noindent\textbf{Quantitative results.} 
We compare our method with existing studies for image generation tasks on various datasets.
The results are summarized in Tab.~\ref{tab:comparison} and Tab.~\ref{tab:comparison_kid}. We can see that for all datasets included, our models consume the smallest MACs while achieving comparable and mostly the best performance. Particularly, we achieve better performance than the original models for almost all datasets while reducing computational cost significantly.
For example, on CycleGAN, our method results in a large compression ratio as the MACs is saved from $56.8$\fu~to $2.56$\fu~($22.2\times$), while at the same time, the model gets better performance as FID is reduced from $61.53$ to $53.48$ for Horse$\shortrightarrow$Zebra. For the Cityscapes dataset with Pix2pix model, we compress the model from $56.8$\fu~ to $5.57$\fu~MACs, which is $10.2\times$ smaller,  while increase the mIoU from $42.06$ to $42.65$. 
Again, for Pix2pix on the Map$\shortrightarrow$Aerial photo dataset, the MACs is reduced from $56.8$\fu~to $4.59$\fu~ by our method, with a compression ratio of $12.4\times$, whereas the FID is improved and reduced from $47.76$ to $45.63$.  

\begin{figure}[t]
\centering
\begin{subfigure}[t]{1\linewidth}
\centering
\includegraphics[width=1\linewidth]{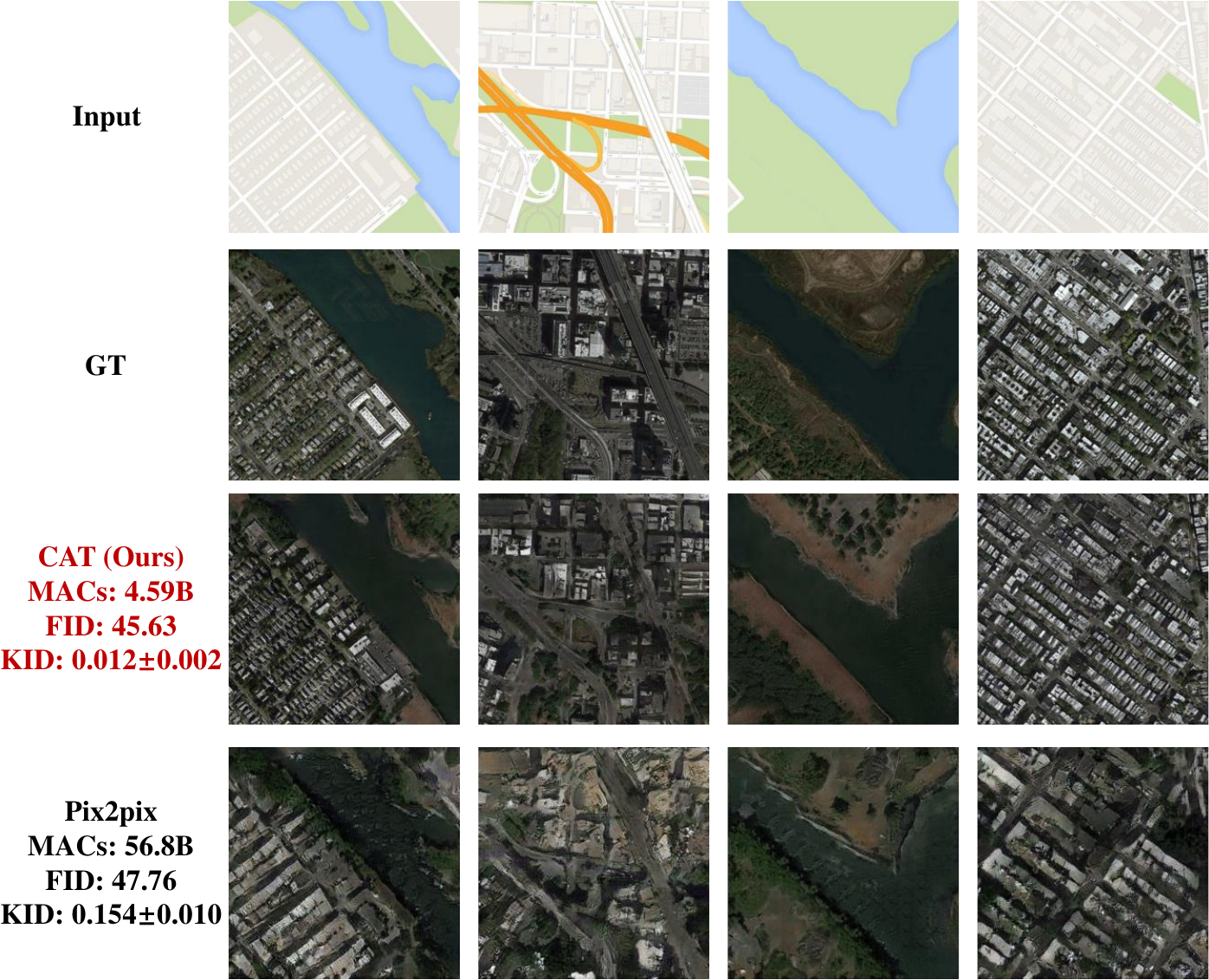}
\label{fig:map}
\end{subfigure}
\\
\begin{subfigure}[t]{1\linewidth}
\centering
\includegraphics[width=1\linewidth]{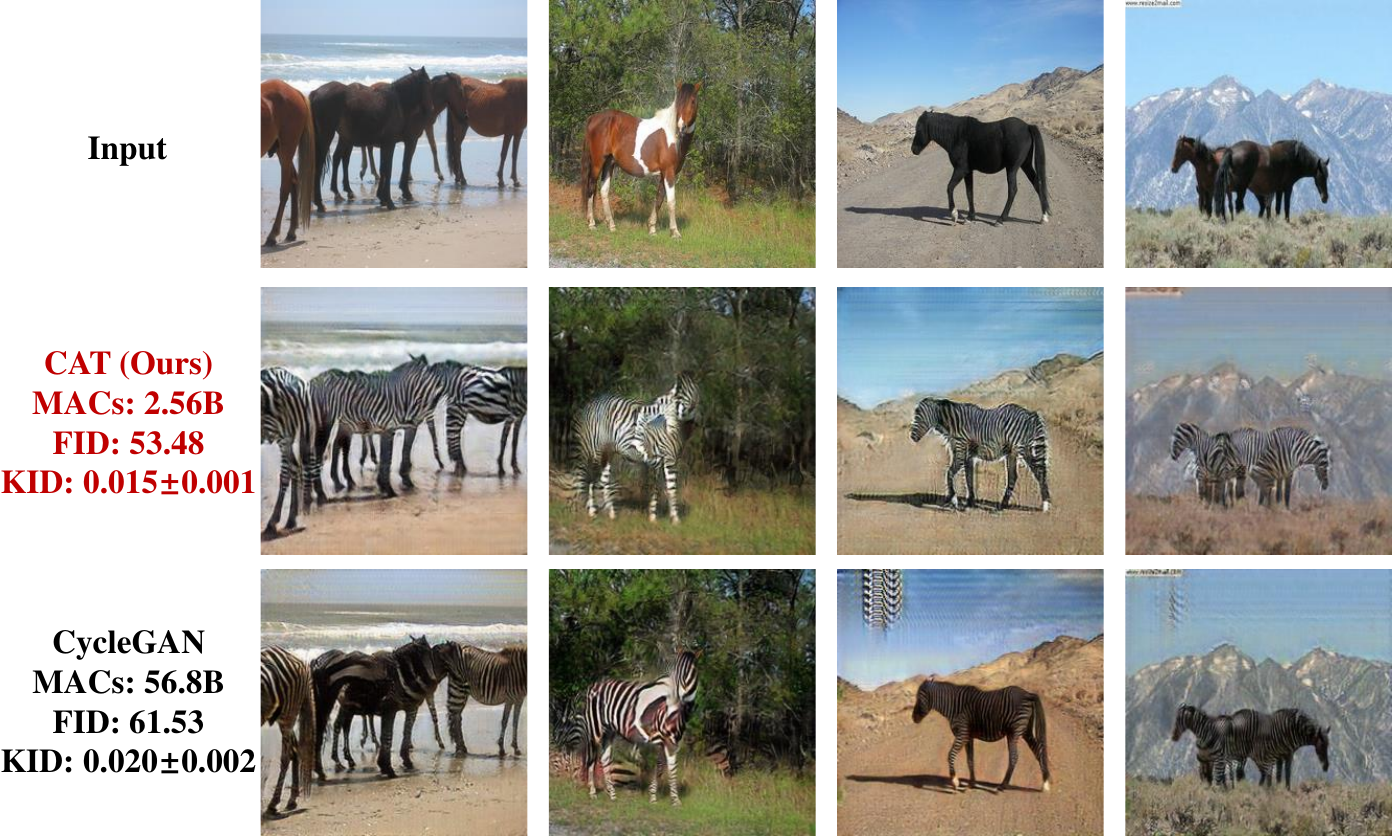}
\label{fig:horse}
\end{subfigure}
\vspace{-15pt}
\caption{Qualitative results on Map$\shortrightarrow$Aerial photo (top four rows) and Horse$\shortrightarrow$Zebra datasets (bottom three rows). 
Compared with original networks (Pix2pix and CycleGAN), our models have much reduced MACs and can generate images with higher fidelity (lower FID) by synthesizing textures that are not well-handled by the original large models.}
\label{fig:map_horse}
\vspace{-15pt}
\end{figure}

To further verify the effectiveness of our method for compressing generative models, we experiment on GauGAN with two target MACs: $30$\fu~and $5.6$\fu.
We choose $5.6$\fu~as it is similar to our compressed Pix2pix model on Cityscapes.
We find that with $30$\fu~MACs, which is $9.4\times$ smaller than GauGAN, the mIoU of our model is better than the original, which is increased from $62.18$ to $62.35$. We further compress the model to less than $5.6$\fu~with a compression ratio of $50.9\times$, and the mIoU is reduced to $54.71$. However, it is still much better than that from the Pix2pix model. These demonstrate that our method is a sound technique for compressing image-to-image models, and provides the state-of-the-art trade-off between computation complexity and image generation performance.

\noindent\textbf{Qualitative results.}
We further show qualitative results to illustrate the effectiveness of our method. Fig.~\ref{fig:city_gaugan} provides samples on Cityspaces, including input segmentation maps, ground-truth (GT), and generated images by different methods. Our compressed model (CAT-A)  achieves better quality (higher mIoU and lower FID) than GauGAN. For example, 
for the leftmost image in Fig.~\ref{fig:city_gaugan}, the back of the car synthesized by CAT-A is clearer than GauGAN, and CAT-A generates less blurry human images than GauGAN for the rightmost image. CAT-B, which has much-reduced MACs than GauGAN ($50.9\times$), can also achieve better image fidelity (lower FID) than GauGAN.
For Map$\shortrightarrow$Aerial photo with Pix2pix (Fig.~\ref{fig:map_horse}), our method generates images with better quality for the river and buildings than the original Pix2pix model.
For Horse$\shortrightarrow$Zebra on CycleGAN,
our method can synthesize better zebra images for challenging input horse images, where the CycleGAN fails to generate.

The examples shown in Fig.~\ref{fig:city_gaugan} \&~\ref{fig:map_horse}  demonstrate that our compression technique is an effective method for saving the computational cost of generative models. Besides, the compressed models can surpass the original models, even though they require much reduced computational cost and, thus, are more efficient during inference. 
These results indicate significant redundancy in the original large generators, and it is worth further studying the extreme of these generative models in terms of performance-efficiency trade-off.

\noindent\textbf{Analysis of searching cost.}
Here we show the analysis of searching costs for finding a student network.
Our method can search the architecture under a pre-defined computational budget with a much reduced searching cost compared with previous state-of-the-art compressing method~\cite{li2020gan}.
Tab.~\ref{tab:searching} provides the searching cost of the two methods on various datasets and models. As can be seen, our method is at least $10,000\times$ times faster for searching. 
The searching time for the previous method~\cite{li2020gan} is estimated by only including the time for training a supernet, which is designed for architecture search.
We estimate it as $20$ hours with $1$ GPU for the CycleGAN and Pix2pix models and $40$ hours with $8$ GPUs for the GauGAN model, both of which are much shorter than those required in practice and thus serves as a lower bound. Besides, we have ignored the time required for searching a student network from the supernet for~\cite{li2020gan},
which is also non-negligible. For example, for Cityscapes with Pix2pix model, the supernet
includes more than $5,000$ possible architectures, and each requires around $3$ minutes with 1 GPU for evaluation, 
resulting in several days of architecture search. Despite, we do not take this process of~\cite{li2020gan} into account for time-estimation in Tab.~\ref{tab:searching}.

\begin{table}[h]
    \centering
    \caption{Architecture search cost, measured in seconds of GPU computation, for our method vs.\ Li \etal~\cite{li2020gan}, across different models.}
    \label{tab:searching}
    \scalebox{0.8}{
    \begin{tabular}{lclc}
        \toprule
        % \multirow{2}{*}{Model} & \multirow{2}{*}{Dataset} &\multirow{2}{*}{Method} & \multirow{2}{*}{MACs} & \multicolumn{2}{c}{Metric} & \multirow{2}{*}{Searching Cost}  \\ 
    % \cmidrule{5-6}
    % & & &  & FID$\downarrow$ & mIoU$\uparrow$ & \\\midrule
    Model & Dataset & Method & \makecell{Search Cost\\(GPU Seconds)} \\ \midrule
        % numbers for cyclegan
        \multirow{4.5}{*}{CycleGAN} & \multirow{2}{*}{Horse$\shortrightarrow$Zebra} & Li \etal~\cite{li2020gan} &  $\gtrsim 7.2\times10^4$ \\
        & & \textbf{CAT (Ours)} & \textbf{3.81}\\
        \cmidrule{2-4}
        & \multirow{2}{*}{Zebra$\shortrightarrow$Horse} & Li \etal~\cite{li2020gan} &  $\gtrsim 7.2\times10^4$ \\
        & & \textbf{CAT (Ours)} & \textbf{3.62}\\
        \midrule
        % ======= numbers for pix2pix edge to shoes
        % \multirow{12}{*}{Pix2pix} & \multirow{4}{*}{edge$\rightarrow$shoes}  & Original~\cite{pix2pix2017} &  56.8\fu  &   24.18 & - & -\\\cmidrule{3-7}
        % & & Li \textit{et al.}~\cite{li2020gan} & 4.81\fu &  26.60 & - & \textcolor{red}{xx\scu}\tnote{$\dagger$}\\
        % & & \textcolor{red}{XXX-A} (\textbf{Ours}) & \textcolor{red}{xx}  & \textcolor{red}{xx} & - & \textcolor{red}{xx\scu}\\\cmidrule{2-7}
        % =======  numbers for pix2pix Cityscapes
        \multirow{4}{*}{Pix2pix}  & \multirow{2}{*}{Cityscapes}  & Li \etal~\cite{li2020gan} & $\gtrsim 7.2\times10^4$\\
        & &  \textbf{CAT (Ours)} & \textbf{4.28}\\\cmidrule{2-4}
        % ======= numbers for pix2pix map
        & \multirow{2}{*}{Map$\shortrightarrow$Aerial photo}  & Li \etal~\cite{li2020gan} & $\gtrsim 7.2\times10^4$\\
        % & & Ours w/ Atomnas & \textcolor{red}{xx} & \textcolor{red}{xx} & - & \textcolor{red}{xx}\\
        & & \textbf{CAT (Ours)} & \textbf{4.33}\\\midrule
        % ======= numbers for gaugan
        \multirow{3}{*}{GauGAN} & \multirow{3}{*}{Cityscapes}  & Li \etal~\cite{li2020gan} & $\gtrsim 1.2\times10^6$\\
        & &  \textbf{CAT-A (Ours)} & \textbf{8.22}\\
        %& & & 28.0\fu & - & 59.78 &\textcolor{red}{xx}\\
        %& & & 20.0\fu & - & 58.54 &\textcolor{red}{xx}\\
        %& & & 15.0\fu & - & 57.88 &\textcolor{red}{xx}\\
        %& & & 10.0\fu & - & 56.13 &\textcolor{red}{xx}\\
        & &  \textbf{CAT-B (Ours)} & \textbf{6.20}
        % \\\midrule
        % % ======= numbers for clade not sure whether this is necessary
        % \multirow{3}{*}{Clade} & \multirow{3}{*}{Cityscapes}  & Li \etal~\cite{li2020gan} & $\gtrsim 1.2\times10^6$\\
        % % \multirow{2}{*}{Clade} & \multirow{2}{*}{Cityscapes}  & \textbf{CAT-A (Ours)} & 7.30~\scu \\
        % & & \textbf{CAT-A (Ours)} & \textbf{7.30} \\
        % & & \textbf{CAT-B (Ours)} & \textbf{4.66}
        \\\bottomrule
    \end{tabular}
    }
\vspace{-10pt}
\end{table}

\section{Conclusion}
In this paper, we study the problem of compressing generative models, especially the generators for image-to-image tasks. 
We show the problem can be tackled by using a powerful teacher model, which is not restricted to teach a student through knowledge distillation, 
but can serve as a supernet to search efficient architecture (for student) under pre-defined computational budgets.

Specifically, our framework is built upon a newly designed teacher model, which incorporates the proposed IncResBlock. 
We show such teacher model contains a large search space where efficient student architecture can be determined through network searching. 
The searching process is implemented with our proposed one-step pruning algorithm, which can be conducted with negligible efforts. 
We also introduce a similarity-based knowledge distillation technique to train student network, where feature similarity between student and teacher is measured directly by the proposed KA index.
With our method, we can obtain networks that have similar or even better performance than original Pix2pix, CycleGAN, and GauGAN models on various datasets. 
More importantly, our networks have much reduced MACs than their original counterparts.

Our work demonstrates that there remains redundancy in existing generative models, and we can achieve improved performance, \eg, synthesizing images with better fidelity, with much reduced computational cost. It is worth further investigating the ability of generative models to synthesize images with high quality under an extremely constrained computational budget, which we leave for future study.
\vspace{-5pt}
\section{Acknowledgement}
The Authors would like to appreciate Jieru Mei from John Hopkins University for invaluable technical discussion. Also this research is partially supported by National Science Foundation CNS-1909172.

{\small
\bibliographystyle{ieee_fullname}
\bibliography{egbib}
}

\appendix

\setcounter{section}{0}
\renewcommand{\thesection}{S\arabic{section}}
\setcounter{table}{0}
\renewcommand{\thetable}{S\arabic{table}}%
\setcounter{figure}{0}
\renewcommand{\thefigure}{S\arabic{figure}}%

\clearpage

{\noindent\large\bf Appendix}

\section{Implementation Details}
In this section, we provide more implementation details in our work.

\begin{table*}[t!]
    \centering
    \caption{Hyper-parameter setting for teacher and student training.}
    \label{tab:hyper}
    \begin{tabular}{cccccccccc}
        \toprule
        \multirow{2.5}{*}{Model} & \multirow{2.5}{*}{Dataset} & \multicolumn{2}{c}{Training Epochs} & \multirow{2.5}{*}{$\lambda_\text{distill}$} & \multirow{2.5}{*}{$\lambda_\text{recon}$} & \multirow{2.5}{*}{$\lambda_\text{fm}$} & \multirow{2.5}{*}{GAN Loss} & ngf & \multirow{2.5}{*}{ndf} \\
        \cmidrule{3-4}\cmidrule{9-9}
        && Const & Decay &&&&& Teacher & \\
        \midrule
        % cyclegan setting
        CycleGAN & Horse$\shortrightarrow$Zebra & 500 & 500 & 0.7 & 5 & - & LSGAN & 64 & 64 \\
        % & Zebra$\shortrightarrow$Horse & 500 & 500 & 0.1 & 5 & - & LSGAN & 64 & 64 \\
        \midrule
        % pix2pix setting
        \multirow{2}{*}{Pix2pix} & Cityscapes & 500 & 750 & 1.0 & 100 & - & Hinge & 64 & 128 \\
        & Map$\shortrightarrow$Aerial photo & 500 & 1000 & 0.8 & 100 & - & Hinge & 64 & 128 %\\
        %& edge$\rightarrow$shoes  & \textcolor{red}{xx}  &\textcolor{red}{xx} & 
        \\\midrule
        GauGAN & Cityscapes & 100 & 100 & 0.5 & 10 & 10 & Hinge & 64 & 64
        \\\bottomrule
    \end{tabular}
\end{table*}

\paragraph{Training details.}
For CycleGAN and Pix2pix models, we use batch size of $32$ for teacher and batch size of $80$ for student, while for GauGAN, the batch size is set to $16$ for both. For each model and each dataset, we apply the same training epochs for teacher and student networks. The learning rate for both generators and discriminators are set as $0.0002$ for all datasets and models.
More detailed training hyper-parameters are summarized in Table~\ref{tab:hyper}.
For the layers used for knowledge distillation between teacher and student networks, we follow the same strategy as Li~\etal~\cite{li2020gan}. Specifically, for Pix2pix and CycleGAN models, the $9$ residual blocks are divided into $3$ groups, each with three consecutive layers, and knowledge is distilled upon the four activations from each end layer of these three groups. For GauGAN models, knowledge distillation is applied on the output activations of $3$ from the total $7$ SPADE blocks, including the first, third and fifth ones.

\paragraph{More details for normalization layers.}
We find that instance normalization~\cite{ulyanov2016instance} without tracking running statistics is critical for dataset Horse$\rightarrow$Zebra
to achieve good performance on the student model, and for dataset Zebra$\rightarrow$Horse, synchronized batch normalization with tracked running statistics gives better performance. For the other datasets batch normalization~\cite{ioffe2015batch} with tracked running statistics is better. Normalization layers without track running statistics introduce extra computation cost, and we take this into account for our calculation of MACs during pruning. Moreover, for GauGAN, we use synchronized batch normalization as suggested by previous work~\cite{park2019semantic,tan2020rethinking}, and  remove the spectral norm~\cite{miyato2018spectral} as we find it does not have much impact on the model performance.

\begin{figure*}[t!]
\centering
\begin{subfigure}[]{.32\textwidth}
\centering
\includegraphics[width=.9\linewidth]{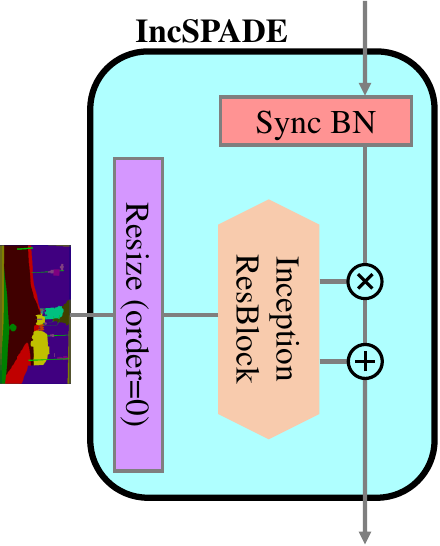}
\label{fig:incspade}
\end{subfigure}
\quad
\begin{subfigure}[]{.42\textwidth}
\centering
\includegraphics[width=.9\linewidth]{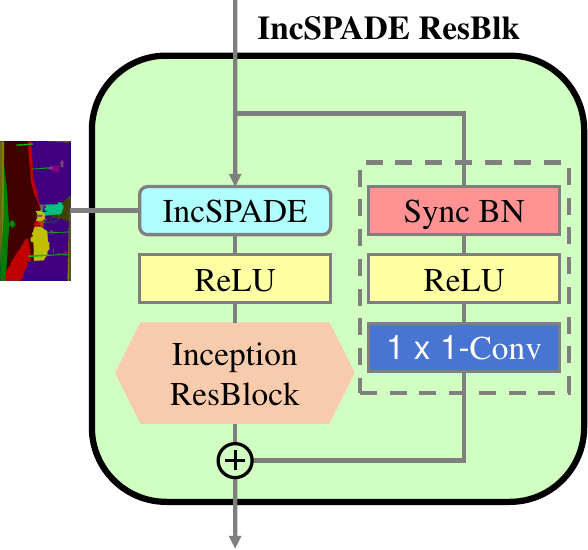}
\label{fig:incresblock}
\end{subfigure}
\caption{SPADE normalization module (IncSPADE, \emph{left}) and SPADE residual block (IncSPADE ResBlk, \emph{right}) with the proposed Inception Resblock (orange hexagon). Note that the optional last normalization layer and residual connection are not applied in the Inception Resblocks that are used in IncSPADE and IncSPADE ResBlk.}
\vspace{-160pt}
\label{fig:spade}
\end{figure*}

% \begin{figure}[t]
% \centering
% \begin{subfigure}[t]{.16\textwidth}
% \centering
% \includegraphics[width=.9\linewidth]{latex/imgs/incspade.pdf}
% \label{fig:incspade}
% \end{subfigure}
% %\quad
% \begin{subfigure}[t]{.24\textwidth}
% \centering
% \includegraphics[width=.9\linewidth]{latex/imgs/incresblock.pdf}
% \label{fig:incresblock}
% \end{subfigure}
% \caption{\todo{Pipeline. Content in the figure: Overall pipeline (pruning, cka), searching space design}}
% \label{fig:spade}
% \end{figure}

\paragraph{Network details for GauGAN.}
For GauGAN, we find it is sufficient for each spade residual block to keep only the first SPADE module in the main body while replace the second one as well as the one in the shortcut by synchronized batch normalization layer. This saves computation cost by a large extent. Besides, we use learnable weights for the second synchronized block for the purpose of pruning. These weights do not introduce extra computation cost, as the running statistics are estimated from training data and not recalculated during inference, enabling fusing normalization layers into the convolution layers. Further, we replace the three convolution layers in the SPADE module by our proposed inception-based residual block (IncResBlock), with normalization layers included for pruning. The details for the architecture are illustrated in Figure~\ref{fig:spade}. We name our SPADE module as IncSPADE and SPADE residual block as IncSPADE ResBlk.

To prune the input channel for each model, we add an extra normalization layer (synchronized batch normalization) with learnable weights after the first fully-connected layer, and prune its channels together with other normalizations using our pruning algorithm described in the Section 3.2 of the main paper. During pruning, we keep the ratio of input channels between different layers as the original model, and the lower bound for the first layer (which has the largest number of channels) is determined by that for the last layer multiplied by the channel ratio, so that all channels are above the bound and the channel ratio is unchanged.

\section{Ablation Analysis of Knowledge Distillation}
Here we show the ablation analysis for knowledge distillation methods. We use our searching method to find a student architecture on Pix2pix task using the Cityscapes dataset, and compare student training without knowledge distillation, with MSE distillation as in~\cite{li2020gan}, and the similarity-based distillation we proposed. The results are summarized in Tab.~\ref{tab:distillation}, where \emph{w/o Distillation} denotes training the student without distillation, and \emph{w/ MSE; Loss Weight $0.5$} and \emph{w/ MSE; Loss Weight $1.0$} denotes MSE distillation with weight $0.5$ and $1.0$, respectively.
We find that distillation indeed improves performance, and our distillation method, which employs KA to maximize feature similarity, is better than MSE on transferring knowledge from teacher to student via intermediate features.

\begin{table}[h!]
    \centering
    \caption{Analysis of knowledge distillation methods on Cityscapes dataset with the Pix2pix setting. Our methods (KDKA) achieves the best result.}
    \label{tab:distillation}
    \begin{tabular}{c|c}
        \toprule
        % Method & MACs & mIoU$\uparrow$  \\ \midrule
        % Atomnas w/o Distillation \\
        % Atomnas w/ MSE \\
        % Atomnas w/ CKA \\ \midrule
        Method & mIoU$\uparrow$ \\ \midrule
         w/o Distillation  & 39.39\\
         w/ MSE; Loss Weight 0.5 & 39.83\\ 
         w/ MSE; Loss Weight 1.0  & 39.76\\
         Ours  & \textbf{42.65}\\\bottomrule
    \end{tabular}
\end{table}

% \begin{table}[h]
%     \centering
%     \caption{Analysis of distillation methods on Cityscape dataset using the Pix2pix model.}
%     \label{tab:distillation}
%     \begin{tabular}{c|c}
%         \toprule
%         % Method & MACs & mIoU$\uparrow$  \\ \midrule
%         % Atomnas w/o Distillation \\
%         % Atomnas w/ MSE \\
%         % Atomnas w/ CKA \\ \midrule
%         Method & mIoU$\uparrow$ \\ \midrule
%          w/o Distillation  & 39.39\\
%          w/ MSE; Loss Weight: 0.5 & 39.83\\ 
%          w/ MSE; Loss Weight: 1.0  & 39.76\\ 
%          w/ CKA  & \textcolor{blue}{38.22} \\ 
%          Ours-centered  & \textcolor{blue}{40.03} \\ 
%          Ours  & 42.53\\\bottomrule
%     \end{tabular}
% \end{table}

\section{More Qualitative Results}
\begin{figure*}[]
\centering
\includegraphics[width=.61\linewidth]{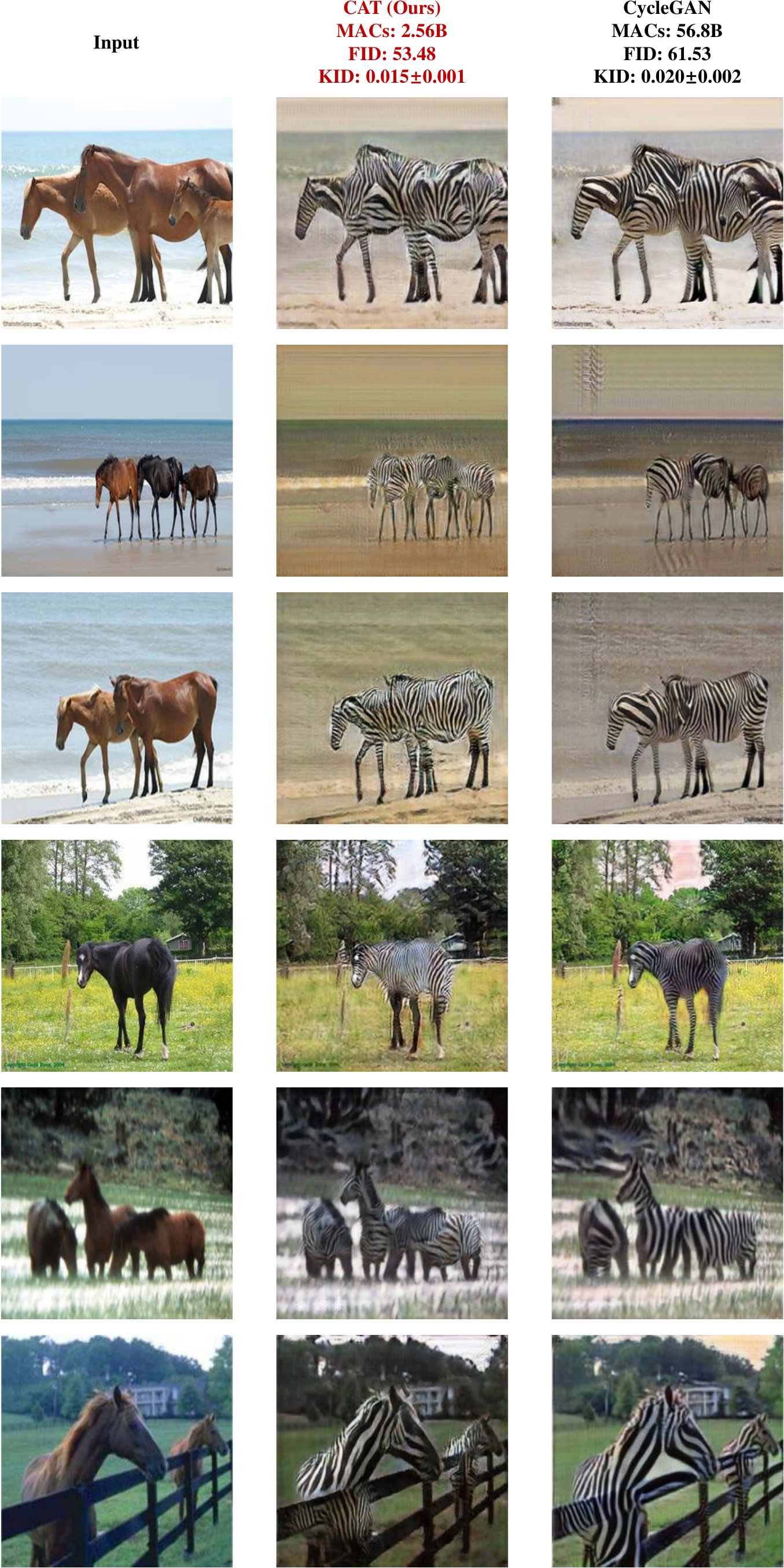}
\caption{
More results on Horse$\shortrightarrow$Zebra dataset. Compared with original CycleGAN, our model has much reduced MACs and can generate images with higher fidelity (lower FID).}
\label{fig:supp_horse}
\end{figure*}

\begin{figure*}[]
\centering
\includegraphics[width=.95\linewidth]{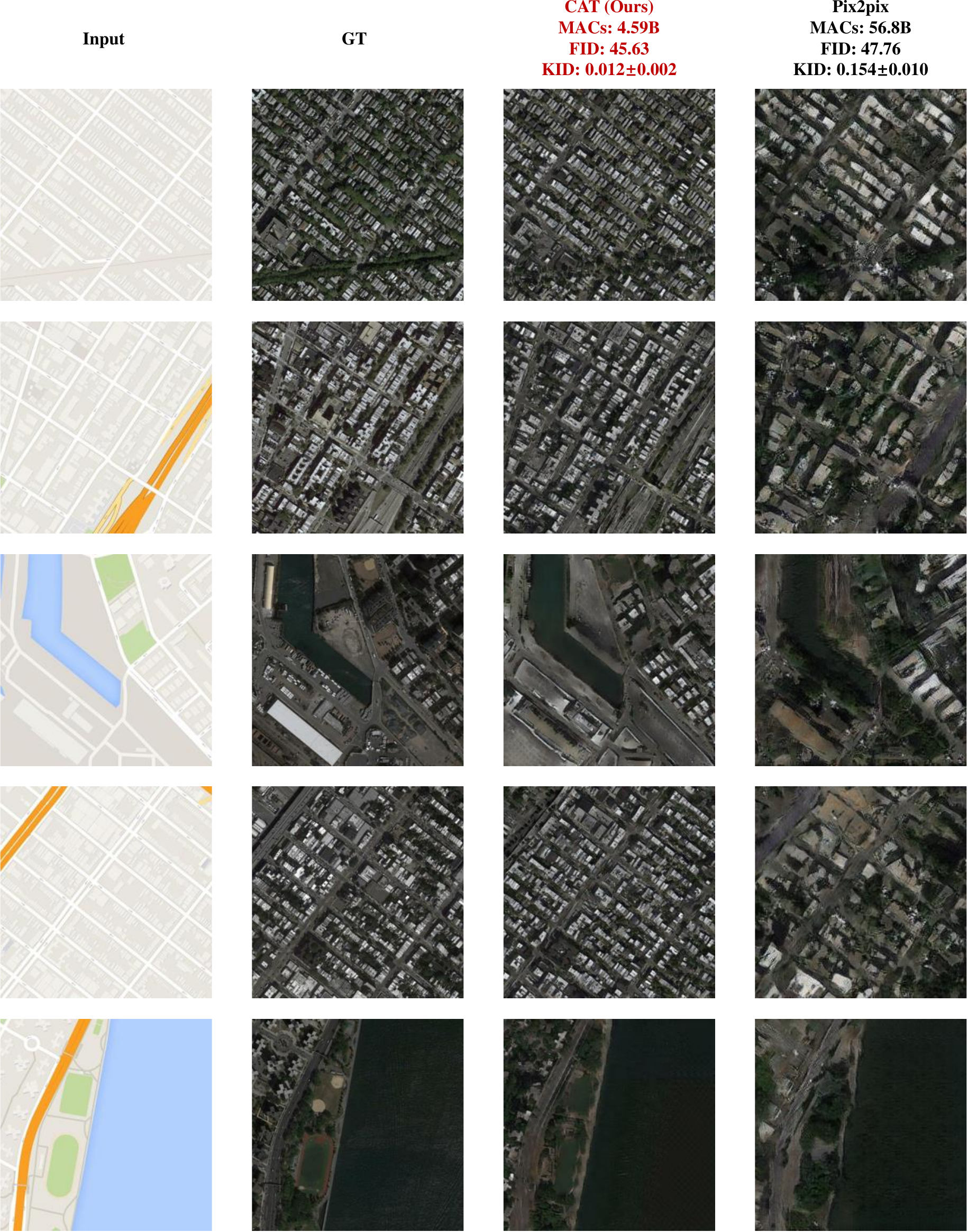}
\caption{
More results on Map$\shortrightarrow$Aerial photo dataset. Compared with original Pix2pix, our model has much reduced MACs and can generate images with higher fidelity (lower FID).}
\label{fig:supp_map}
\end{figure*}

\begin{figure*}[h!]
\centering
\includegraphics[width=\linewidth]{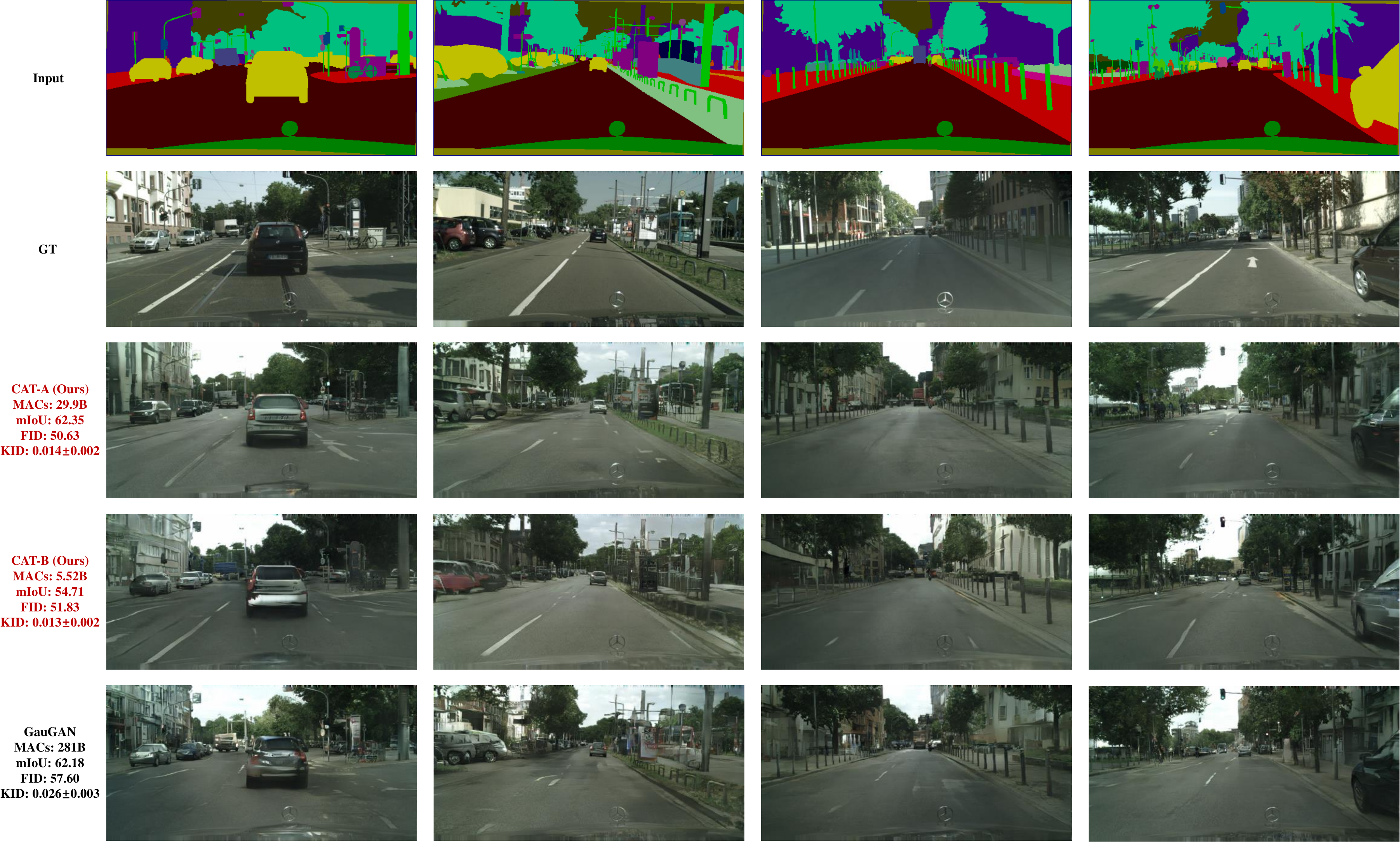}
\caption{
More qualitative results on Cityscapes dataset. 
Images generated by our compressed model (CAT-A, third row) have higher mIoU and lower FID than the original GauGAN model (fifth row), even with much reduced computational cost.
For our CAT-B model (fourth row, $50.9\times$ compressed than GauGAN), although it has lower mIoU, the CAT-B model can synthesize higher fidelity images (lower FID) than GauGAN.}
\vspace*{4in}
\label{fig:supp_spade_city}
\end{figure*}

We show more qualitative results for CycleGAN on Horse$\shortrightarrow$Zebra, Pix2pix on Map$\shortrightarrow$Aerial photo, as well as GauGAN on Cityscapes in Figs.~\ref{fig:supp_horse},~\ref{fig:supp_map}, and~\ref{fig:supp_spade_city}, respectively.

\end{document}